\documentclass[runningheads]{llncs}

\usepackage{eccv}

\usepackage{graphicx}
\newcommand{\safeincludegraphics}[2][]{%
  \IfFileExists{#2}{\includegraphics[#1]{#2}}{%
    \fbox{\begin{minipage}[c][0.28\linewidth][c]{0.9\linewidth}\centering Missing figure: \texttt{#2}\end{minipage}}%
  }%
}
\usepackage{amsmath}
\usepackage{amssymb}
\usepackage{booktabs}
\usepackage{makecell}
\usepackage{enumitem}
\usepackage{multirow}
\usepackage{pifont}
\usepackage{tabularx}

\newcommand{\method}{\text{AgentMV}}
\newcommand{\cqr}{\mathrm{CQR}}

\usepackage{hyperref}
\title{AgentMV: A State-Guided Multi-Agent Framework for Budget-Aware Music Video Generation}
\titlerunning{AgentMV}

\author{Huimin Wang \quad Chang Xia \quad Leilei Ouyang \\ Yongqi Kang \quad Yu Fu \quad Yuqi Ouyang\thanks{Corresponding author: \texttt{yuqi.ouyang@scu.edu.cn}}}
\authorrunning{Huimin Wang et al.}
\institute{College of Computer Science, Sichuan University, Chengdu}

\begin{document}
\maketitle

% ================================================================
%  Abstract
% ================================================================
\begin{abstract}
Generating a complete music video from a song requires more than synthesizing visually plausible clips for individual lyric prompts. A practical system must maintain long-range visual consistency, coordinate recurring motifs, synchronize edits with musical structure, and manage the cumulative cost of video generation. Existing approaches typically generate segments independently or adopt fixed generation strategies, limiting their ability to perform global planning over an entire song. We present \method{}, a state-guided multi-agent framework for budget-aware music video generation. \method{} decomposes the production process into specialized agents for music perception, script planning, visual asset provision, segment realization, and final assembly, coordinated through a Structured Persistent State that enables information exchange and state tracking throughout the generation process. To optimize generation resources, we formulate motif-aware segment realization as a group-level Multiple-Choice Knapsack Problem solved via dynamic programming, considering segment importance, generation quality, cost, and motif reuse under a global budget constraint. Experiments on song benchmarks demonstrate that \method{} improves quality-cost trade-offs over existing MV generation frameworks, highlighting the potential of state-guided multi-agent coordination and budget-aware planning for long-form music video generation with improved quality and efficiency.

\end{abstract}

% ================================================================
%  1. Introduction
% ================================================================
\section{Introduction}
\label{sec:intro}

\begin{figure}[!t]
  \centering
  \safeincludegraphics[width=0.85\linewidth]{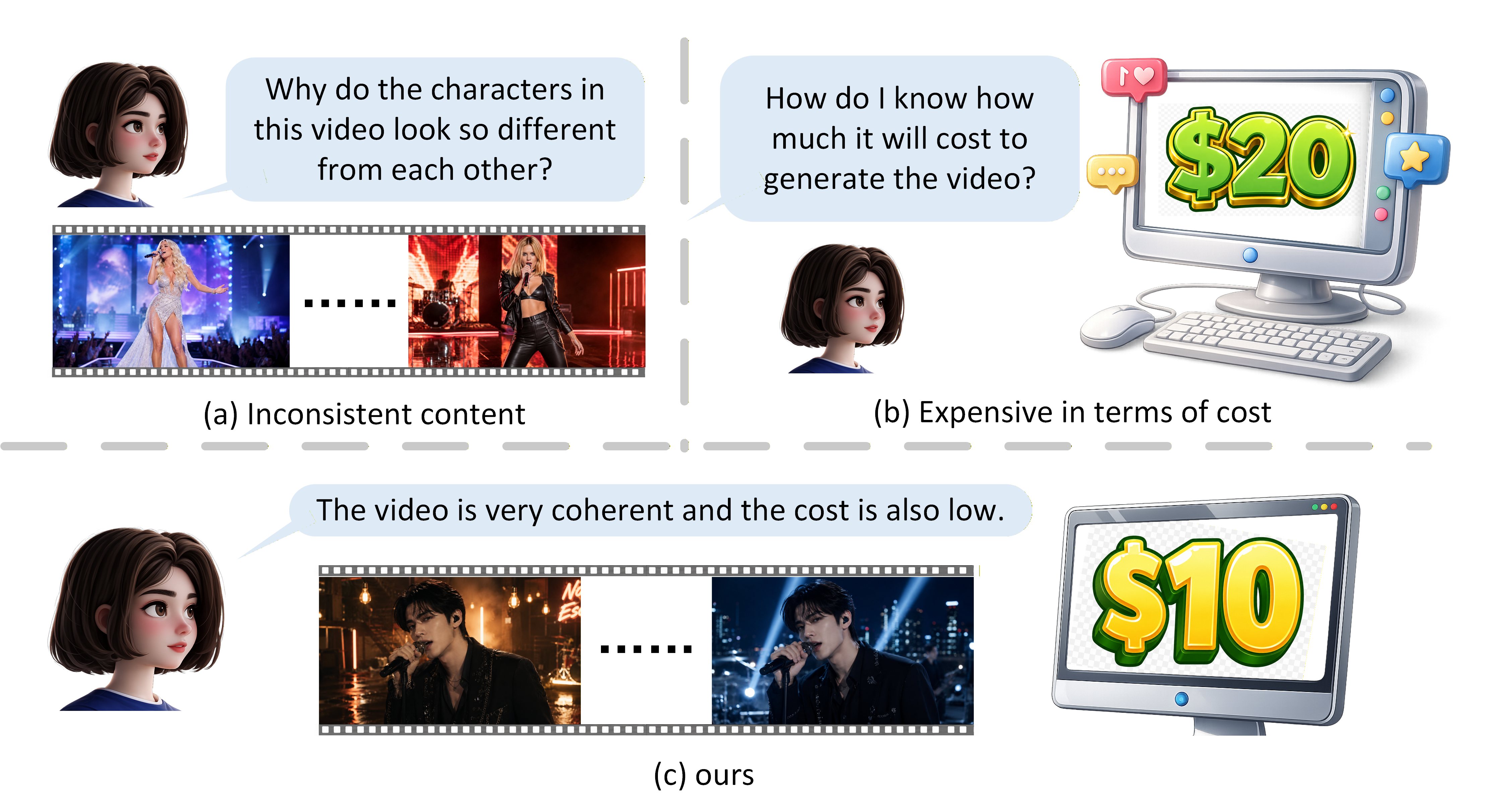}
  \caption{Motivation for budget-aware, long-form MV generation. Independently generated clips may exhibit inconsistent characters and scenes, while uniformly applying expensive generation settings provides no explicit control over total cost. Our framework addresses these limitations through state-guided coordination and budget-aware generation, improving long-range visual consistency while reducing the cost.}
  \label{fig:limit}
\end{figure}

Recent diffusion models can generate visually compelling short video clips, yet full-song music video (MV) generation remains challenging. Unlike standalone clips, an MV must jointly maintain long-range visual coherence, motif recurrence, edit synchronization with musical timing, and cost-efficient generation. As illustrated in \cref{fig:limit}, existing MV generation pipelines often generate segments independently or apply uniform generation policies across an entire song, leading to inconsistent visual identities and limited control over cumulative cost. Although recent long-video generation methods improve temporal consistency or generation length~\cite{Wang2023GenLVideo,Qiu2024FreeNoise,Zhou2024StoryDiffusion}, they do not explicitly address the structured planning challenges of complete-song MV generation, where visual coherence, rhythmic synchronization, and generation efficiency must be optimized jointly.

Existing MV generation systems introduce a second limitation. Methods such as AutoMV~\cite{Tang2025AutoMV}, MuseV~\cite{musev2024}, and VideoComposer~\cite{Wang2024videocomposer} largely follow fixed generation policies that treat song segments independently or allocate generation resources uniformly. Such strategies ignore the unequal importance of different musical sections and repeatedly regenerate visually related content across recurring motifs, increasing synthesis cost while providing limited benefit to narrative continuity. Conversely, reducing generation quality independently for selected segments can introduce identity drift and inconsistent visual structure. As a result, these approaches lack an explicit mechanism for balancing generation quality, long-range coherence, and production cost over an entire song.

% This design makes dependencies explicit: scripts refer to persistent characters and scenes, motif consumers refer to reusable owner assets, and realized clips remain linked to their assigned actions.

% During realization, recurring motifs share a visual prefix and generate only divergent suffixes, reducing redundant synthesis while preserving motif identity.

\begin{figure}[!t]
  \centering
  \safeincludegraphics[width=0.98\linewidth]{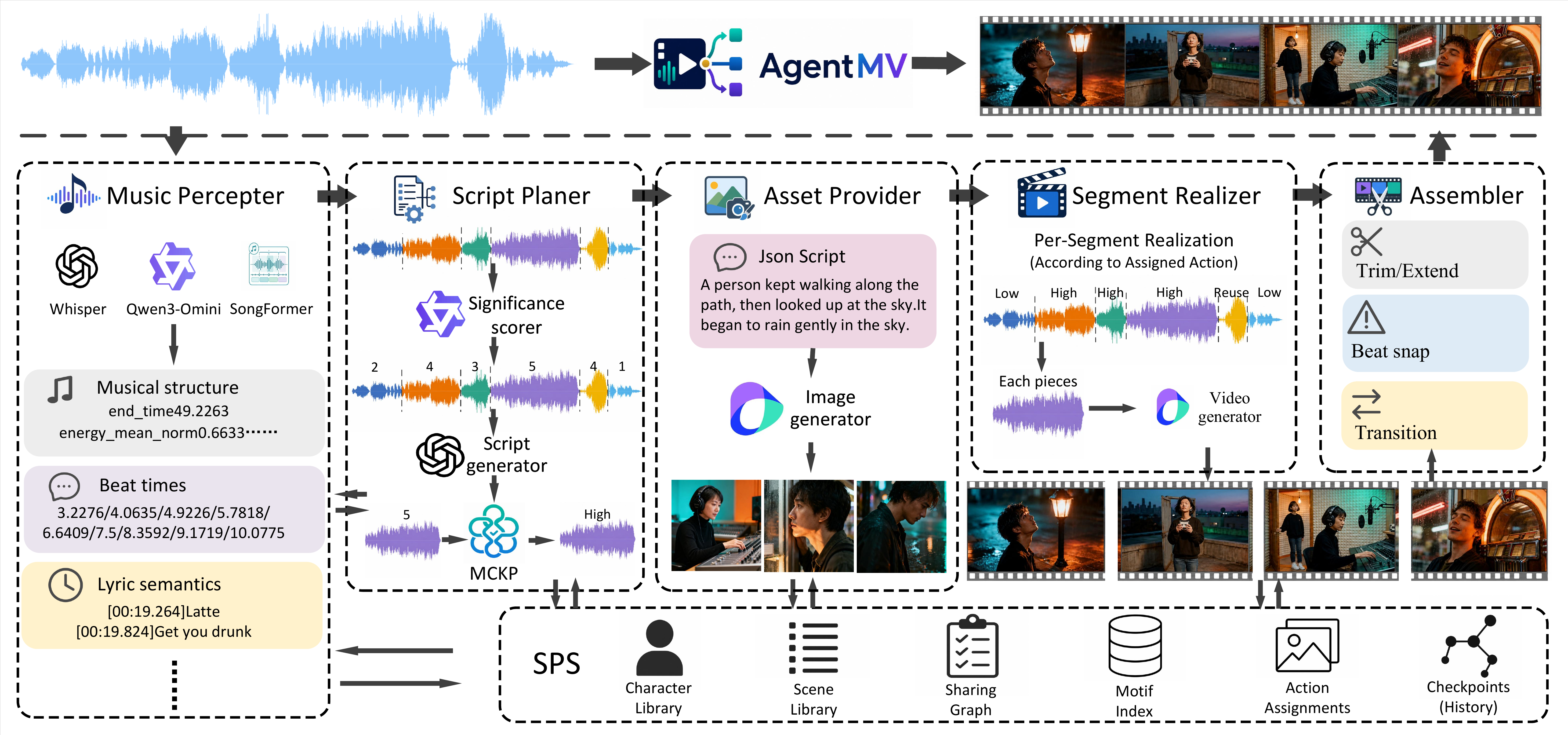}
  \caption{Overview of \method{}. The framework decomposes long-form music video generation into music perception, script planning, visual asset provision, segment realization, and final assembly. A Structured Persistent State (SPS) maintains the shared production state across agents, while a group-level MCKP performs budget-aware allocation before video synthesis.}
  \label{fig:overview}
\end{figure}

These observations suggest three requirements for budget-aware MV generation: a structured representation of global creative intent, a planning mechanism that accounts for non-local motif dependencies, and an execution pipeline that preserves planning decisions through final assembly. We address them with \method, a five-agent system comprising Music Percepter, Script Planner, Asset Provider, Segment Realizer, and Assembler. Depicted in \cref{fig:overview}, rather than exchanging unconstrained dialogue, the agents communicate through Structured Persistent State (SPS) that stores typed intermediate artifacts and is checkpointed after each stage. To allocate resources effectively, the Script Planner estimates segment saliency, constructs motif-sharing groups, and formulates plan selection as a group-level Multiple-Choice Knapsack Problem (MCKP)~\cite{Sinha1979MCKP}, which is solved via dynamic programming to prioritize salient segments while reusing shared visual structure across recurring motifs under the global budget. \method addresses generation quality, visual consistency, rhythmic synchronization, and cost through coordinated planning and execution, providing a practical solution for complete-song MV generation. Our main contributions are:

\begin{itemize}[noitemsep,topsep=2pt]
\item We design a SPS-guided five-agent MV generation framework that decomposes long-horizon production into perception, script planning, asset creation, segment realization, and assembly, enabling explicit stage-wise coordination of content generation and temporal execution.
\item We formulate motif-aware budget allocation as a group-level Multiple-Choice Knapsack Problem (MCKP) and develop a dynamic-programming solver for optimal plan selection under the formulated budget objective, enabling cost-effective long-horizon MV generation through coordinated motif reuse and quality allocation.
\item Experiments on a five-song pilot benchmark demonstrate that \method{} achieves a stronger quality--cost trade-off than existing MV generation frameworks, obtaining the best BeatAlign, CQR, and motif consistency with the lowest measured generation cost while remaining competitive in visual-semantic alignment.

\end{itemize}

% \begin{table}[t]
% \centering
% \footnotesize
% \setlength{\tabcolsep}{2.2pt}
% \caption{Capability comparison of representative music-video generation systems.}
% \label{tab:capability_comparison}
% \begin{tabular}{lccccc}
% \toprule
% \textbf{Method}
% & \makecell{\textbf{MV}\\\textbf{oriented}}
% & \makecell{\textbf{Music}\\\textbf{aware}}
% & \makecell{\textbf{Full}\\\textbf{song}}
% & \makecell{\textbf{Motif}\\\textbf{memory}}
% & \makecell{\textbf{Budget}\\\textbf{reuse}} \\
% \midrule
% Music2Video~\cite{Kim2022Music2Video}
% & \cmark & \cmark & \xmark & \xmark & \xmark \\
% MV-Crafter~\cite{MVCrafter}
% & \cmark & \cmark & \xmark & \xmark & \xmark \\
% AI-authored pipeline~\cite{AIMVPipeline}
% & \cmark & \cmark & \xmark & \xmark & \xmark \\
% MuseV~\cite{musev2024}
% & \xmark & \xmark & \cmark & \xmark & \xmark \\
% AutoMV~\cite{Tang2025AutoMV}
% & \cmark & \cmark & \cmark & \cmark & \xmark \\
% \midrule
% \textbf{\method{}}
% & \cmark & \cmark & \cmark & \cmark & \cmark \\
% \bottomrule
% \end{tabular}
% \vspace{-0.5em}
% \end{table}

% ================================================================
%  2. Related Work
% ================================================================
\section{Related Work}
\label{sec:related}

\noindent \textbf{Long-horizon and music-conditioned video.}
Long-video diffusion methods extend temporal coverage through overlapping denoising windows~\cite{Wang2023GenLVideo}, noise rescheduling~\cite{Qiu2024FreeNoise}, or consistency-aware attention mechanisms~\cite{Zhou2024StoryDiffusion}. Music-conditioned work has studied beat-driven motion~\cite{Kim2022Music2Video}, controllable video synthesis~\cite{Wang2024videocomposer}, virtual-human generation~\cite{musev2024}, and automatic full-MV production~\cite{Tang2025AutoMV}. These approaches improve generation quality or temporal extent, but lack globally coordinated planning under resource constraints. In contrast, our framework enables globally coordinated long-horizon MV generation with budget-aware resource optimization.

\medskip
\noindent \textbf{Structured Multi-Agent Coordination.}
Recent agent systems show that long-horizon tasks benefit from explicit roles and structured intermediate representations. MetaGPT adopts SOP-based document exchange~\cite{Hong2024MetaGPT}, while other agentic systems combine hierarchical planning with persistent memory and constrained execution~\cite{Agashe2025AgentS}, model task dependencies via graphs~\cite{Hong2025DataInterpreter}, or adapt collaboration structures to task complexity~\cite{Kim2024MDAgents}. However, these approaches are not tailored to multimedia generation. In contrast, our approach introduces SPS, a media-aware persistent state that bridges agent collaboration and long-horizon MV generation.

\medskip
\noindent \textbf{Resource Allocation under Constraints.}
Resource allocation has been widely studied as constrained optimization, where knapsack variants provide principled budgeted selection and MCKP~\cite{Sinha1979MCKP} extends this formulation to grouped decisions. Related studies further investigate budget-aware model selection and inference-time resource allocation through adaptive configuration strategies~\cite{Xu2018AdaScale,Wang2018BlockDrop}, while recent generative models explore quality-efficiency trade-offs via adaptive inference and diffusion acceleration~\cite{Kahatapitiya2025AdaCache}. However, these approaches optimize resources within individual models or inference trajectories rather than across long-horizon generation workflows. In contrast, our framework performs global budget-aware optimization over the entire MV generation process with coordinated segment-level decisions.

% ================================================================
%  3. Method
% ================================================================
\section{Method}
\label{sec:system}

\subsection{Structured Persistent State Library}
\label{sec:sps_library}
Given audio $A$, lyrics $L$, and a generation budget $B$, \method{} generates a full-length music video through a sequential agentic process involving the Music Percepter, Script Planner, Asset Provider, Segment Realizer, and Assembler. The Music Percepter writes the musical timeline, the Script Planner writes scripts, dependencies, and actions, the Asset Provider writes visual references, the Segment Realizer writes clips and execution records, and the Assembler writes the final video and editing decisions. Each agent reads and updates the state entries associated with its assigned production stage, forming a sequential agentic process for long-horizon music video creation.

Instead of exchanging unconstrained messages, these agents communicate through the SPS shown in the bottom-right block of \cref{fig:overview}. SPS provides a persistent media-centric workspace that maintains visual references, segment dependencies, generation policies, execution history, and intermediate results throughout the MV creation process. The SPS maintains a Character Library and Scene Library for persistent visual references, a directed Sharing Graph for recurring segment dependencies, a Motif Index for reusable content, Action Assignments for generation policies, and Checkpoints for execution history. It also records perceptual timelines, scripts, saliency scores, predicted and realized costs, clip paths, and failure states. Each update to the SPS is committed only when its required fields are present and internally consistent. Failed updates preserve the state at the preceding checkpoint, enabling inspection and recovery. SPS therefore serves simultaneously as the inter-agent protocol, the persistent memory of the MV, and the execution ledger linking the input song to the final output.

\subsection{Music Percepter}
\label{sec:music_percepter}
The Music Percepter writes the musical timeline by converting audio and lyrics $(A,L)$ into the common temporal coordinate system. Whisper provides word-level lyric timestamps~\cite{Radford2022Whisper}, SongFormer estimates structural sections and downbeats~\cite{Hao2025SongFormer}, while Qwen3-Omni extracts acoustic descriptors such as normalized energy~\cite{Xu2025Qwen3Omni}. The agent aligns these signals and partitions the song into $N$ contiguous segments $\mathcal{X}=\{x_i\}_{i=1}^{N}$, with each segment statistics presented as:

\begin{equation}
    x_i=(t_i^s,t_i^e,d_i,r_i,L_i,\mathcal{B}_i,e_i),
\end{equation}
where, $t_i^s$ and $t_i^e$ are the temporal boundaries, $d_i=t_i^e-t_i^s$ is the duration, $r_i$ is the structural role (e.g., verse or chorus), $L_i$ is the aligned lyric span, $\mathcal{B}_i$ contains beat anchors, and $e_i$ is the normalized energy descriptor. This representation is written to the SPS as the perception layer. The Planner next uses the percepted results $(r_i,L_i,e_i,d_i)$ to construct scripts and allocate budget, while the Assembler later uses $(t_i^s,t_i^e,\mathcal{B}_i)$ to recover the intended song timing.

\subsection{Script Planner and Motif-Aware Budget Allocation}
\label{sec:script_planner}
The Script Planner maps the percepted results and budget $B$ to an executable production plan. First, it generates a segment-level script containing the scene, characters, action, cinematography, and continuity constraints. Then, a frozen significance scorer assigns each segment a saliency score $s_i\in\{1,\ldots,5\}$~\cite{Yang2025Qwen25}, representing its production importance. Next, the Planner analyzes structural recurrence and script-level visual concepts to identify reusable motifs. It designates one canonical occurrence as the owner and links later occurrences as consumers in the Sharing Graph. The owner defines the shared character, scene, and motion premise, while each consumer retains a divergent suffix reflecting its local lyrics and musical context.

Let $\mathcal{G}$ denote the set of sharing groups, where each group $g\in\mathcal{G}$ contains one or more segments, and let $P_g$ denote the feasible realization plans for group $g$. A singleton group contains a single segment and admits either a High or Mid action, corresponding to two calibrated quality-cost settings of the video backend. For a sharing group, each plan specifies the quality tier of the shared prefix together with that of every member-specific suffix, while consumers execute the Reuse action by retrieving the owner's generated prefix. With plan cost $c_{gp}$, utility $U_{gp}$, and binary decision variable $y_{gp}$ indicating whether plan $p$ is selected for group $g$, budget allocation is formulated as the following group-level MCKP~\cite{Sinha1979MCKP}:

\begin{align}
  \max_{\{y_{gp}\}}\quad
    &\sum_{g\in\mathcal{G}}\sum_{p\in P_g}U_{gp}y_{gp}
    \label{eq:mckp}\\
  \mathrm{s.t.}\quad
    &\sum_{g\in\mathcal{G}}\sum_{p\in P_g}c_{gp}y_{gp}\leq B,
    \qquad
    \sum_{p\in P_g}y_{gp}=1,\ \forall g\in\mathcal{G}.
    \notag
\end{align}

Let $C(a,d)$ denote the estimated generation cost of applying action $a$ to a segment of duration $d$, and let $Q(a)$ denote the calibrated quality score of action $a$. For a singleton group $g={k}$ with duration $d_k$ and saliency $s_k$, the cost and utility are:

\begin{align}
c_{gp}
&=
C(a_k,d_k), \label{eq:singleton_cost}\\
U_{gp}
&=
s_k \,d_k\, Q(a_k). \label{eq:singleton_utility}
\end{align}

For a sharing group, let $d_i$ denote the shared-prefix duration generated by the owner segment, and $d_j$ denote the unique suffix duration of of consumer segment $j\in g$. A joint plan specifies action $a_i$ for the shared prefix and action $a_j$ for each member-specific suffix, yielding:

\begin{align}
c_{gp}
&=
C(a_i,d_i)+\sum_{j\in g}C(a_j,d_j),
\label{eq:group_cost}\\
U_{gp}
&=
Q(a_i)\sum_{j\in g}s_jd_i
+\sum_{j\in g}Q(a_j)s_jd_j.
\label{eq:group_utility}
\end{align}

The duration terms weight utility by perceptual exposure, while saliency emphasizes visually important segments. Unlike the singleton formulation, \cref{eq:group_cost,eq:group_utility} explicitly represent motif-linked segments through grouped realization plans, enabling motif-aware budget allocation by accounting for a shared-prefix cost once while accumulating its utility across the sharing group. We discretize the budget into integer units and solve \cref{eq:mckp} using dynamic programming, yielding the globally optimal feasible plan under the specified budget. The resulting scripts, saliency scores, owner-consumer relations, prompts, action assignments, and predicted costs are committed to the SPS, forming a binding execution contract for the Asset Provider and Segment Realizer.

\subsection{Asset Provider}
\label{sec:asset_provider}
The Asset Provider converts the symbolic production plan into visual conditioning assets before video synthesis. It reads the character, scene, style, and shot descriptions in the scripts and uses Seedream~\cite{Gao2025Seedream3} to create three asset types: canonical character portraits, canonical scene references, and segment keyframes. A character or scene is first generated, assigned with a persistent identifier, and stored in the corresponding SPS. Segment keyframes are then generated by conditioning on these canonical references, ensuring that non-adjacent segments referring to the same entity share consistent visual anchors. For each sharing group, the owner and its consumers receive compatible keyframes derived from their shared motif specification. The Asset Provider writes the asset identifiers, locations, provenance, and associated generation costs to the SPS. This process transforms textual continuity constraints into explicit visual conditions, reducing the likelihood of inconsistent character, scene, and style realization across segments..

\subsection{Segment Realizer}
\label{sec:segment_realizer}
The Segment Realizer executes the committed action assignments using the scripts and visual assets prepared by the preceding agents. Independent segments are synthesized directly according to their assigned quality tier. For sharing groups, execution follows dependency order, where each owner is realized before its consumers. The owner generates the shared prefix of duration $d_i$ once and registers its reusable clip range in the motif index. Consumers then retrieve this prefix and invoke the backend only for their divergent suffixes of duration $d_j$, producing complete segments through concatenation. Consequently, the realized group cost consists of one shared-prefix generation and multiple suffix generations, consistent with \cref{eq:group_cost}, rather than one full generation per occurrence. Both the shared-prefix and divergent-suffix generation are conditioned on the same canonical character and scene references together with the shared motif specification to preserve visual continuity across the prefix-suffix boundary. The realization process for segment $i$ is presented as:

\begin{equation}
    v_i=\mathcal{F}(x_i,a_i,T_{i},R_i,E_i,M_i),
\end{equation}
where $x_i$ denotes the percepted segment statistics; $a_i$ is the assigned action; $T_i$ is the shot prompt; $R_i$ and $E_i$ are the retrieved character and scene conditions; and $M_i$ is an optional reusable motif entry. Under the assigned \textit{High} or \textit{Mid} quality tier, the Seedance backend~\cite{Gao2025Seedance} synthesizes the corresponding video content.

After each realization, the Segment Realizer records the executed action, prompt and asset references, clip path, duration, reusable range, realized cost, and execution status in the SPS. Missing clips, invalid durations, or unresolved owner references mark failed validation and are not committed. Failed executions preserve the previous checkpoint and enable targeted retry. The SPS therefore maintains the correspondence between the optimized production plan and the generated media.

\subsection{Assembler Agent}
\label{sec:assembler}
The Assembler maps the realized clip set $\{v_i\}_{i=1}^{N}$ back to the original song timeline. It first verifies that all segments have valid clips and preserves the ordering defined by $\mathcal{X}$. Each clip is then trimmed or extended to its planned duration $d_i$. Internal boundaries are aligned to the nearest beat anchor within a predefined tolerance; otherwise, the original boundary is retained. Transition types are selected according to adjacent structural roles, where hard cuts emphasize major section changes and fades or crossfades provide smoother continuity. The resulting clips are concatenated and multiplexed with the original audio to produce the final music video $V$. Since assembly edits existing media without additional video synthesis, these operations incur no extra generation cost. The final SPS checkpoint records $V$, segment ordering, duration adjustments, boundary alignments, transition decisions, and the aggregate cost ledger, completing a traceable mapping from $(A,L,B)$ to a music video $V$.

% \noindent
% We organize the experiments around four questions. 
% \emph{(Q1)} Compared with existing MV generation pipelines, does \method{} improve not only visual-audio alignment but also the practical quality--cost trade-off and controllability of long-form generation? 
% \emph{(Q2)} On complete songs, does structured resource allocation outperform uniform or heuristic generation policies under realistic budget constraints? 
% \emph{(Q3)} Which components of \method{} are responsible for the observed gains, including allocation, beat-synchronous assembly, and motif-aware reuse? 
% \emph{(Q4)} Are the planning signals sufficiently calibrated, and is the optimization layer computationally practical for the planning sizes considered in this work?

% ================================================================
%  4. Experiments
% ================================================================
\section{Experiments}
\label{sec:experiments}

\noindent \textbf{Data.}
Experiments are conducted under two evaluation settings. All generated video clips have a fixed duration of 5 seconds with a height $1080$ and width $1920$. For comparison with prior MV generation systems, we use a controlled five-song excerpt benchmark where each excerpt retains representative musical structure. All methods are evaluated on identical audio excerpts, lyrics, SongFormer-derived segmentation, and random seeds. For allocation analysis and ablations, we use selection of five songs to evaluate long-horizon generation under recurring visual motifs, cumulative generation costs, and complete musical structure. Unless otherwise specified, results are reported as mean and standard deviation over the five songs. 

\medskip
\noindent \textbf{Implementation details.}
Word-level lyric timestamps are obtained using Whisper~\cite{Radford2022Whisper}. SongFormer~\cite{Hao2025SongFormer} provides structural segmentation and downbeat detection, while Qwen3-Omni~\cite{Xu2025Qwen3Omni} estimates acoustic energy and Qwen-Plus~\cite{Yang2025Qwen25} predicts segment saliency. Video generation uses Seedance~\cite{Gao2025Seedance} with Seedream-generated identity and scene images as SPS anchors~\cite{Gao2025Seedream3}. The group-level MCKP is solved by dynamic programming with planning proxies $Q(\mathrm{High})=1.0$ and $Q(\mathrm{Mid})=0.7$. A reused prefix inherits the High or Mid quality setting of its owner. Unless otherwise stated, ablation studies freeze the perception outputs, scripts, scene anchors, and random seeds. The planner uses an estimated generation cost for budget allocation, whereas all reported costs are realized end-to-end monetary costs associated with both video synthesis and auxiliary model calls. For the full-song benchmark, costs are accumulated over the complete generated song to avoid mixing different evaluation scales.

\medskip
\noindent \textbf{Evaluation Metrics.}
Following AutoMV-Bench~\cite{Tang2025AutoMV}, we use a frozen GPT-4o judge to evaluate generation quality at the video level and assess music-content, post-production, and artistic criteria at the segment level. We additionally report computed metrics, including CLIP-Score~\cite{Hessel2021CLIPScore} for visual-semantic alignment between generated visual contents and lyrics; BeatAlign~\cite{Li2021AIChoreographer} for rhythmic synchronization between video cut points and detected downbeats; Motif for motif Consistency, computed as mean SSIM ~\cite{Wang2004SSIM} and LPIPS~\cite{Zhang2018LPIPS} over annotated motif pairs. Additionally, Cost-Quality Ratio (CQR) is introduced for generation efficiency, computed as follows:
\begin{equation}
\cqr=\sum_{i=1}^{N}s_iU_i \big/C,
\end{equation}
where $s_i\in\{1,\ldots,5\}$ is the segment saliency score, $U_{i}\in[1,5]$ denotes segment generation quality estimated by the GPT-4o, and $C$ is the total monetary cost in USD. Since all compared methods generate videos with identical segment durations in each benchmark, CQR provides a normalized measure of saliency-weighted generation quality achieved per unit cost, where higher values indicate better quality-cost efficiency.

\subsection{Performance Comparisons with Previous Methods}
\label{sec:sota}

We compare \method{} with representative MV generation methods on the controlled five-song pilot benchmark. As shown in \cref{tab:baseline_comparison}, \method{} achieves the best BeatAlign, CQR, motif consistency, and generation cost among all compared methods. Compared with AutoMV, \method{} improves CQR from $0.4702$ to $0.7596$ while reducing the average cost from $3.25$ to $2.19$ USD. The substantial BeatAlign improvement demonstrates our strong effectiveness on rhythmic synchronization, improving temporal coherence during the generation process. For visual-semantic alignment, AutoMV the highest CLIP score, while \method{} remains competitive, reflecting the trade-off between local visual-semantic alignment and long-horizon consistency, as \method{} preserves established characters, scenes, and motifs through reuse, which may slightly reduce CLIP but improves motif consistency and cost efficiency. Overall, these findings validate the effectiveness of the proposed budget-aware planning, motif-aware reuse, and beat-synchronous assembly strategies for achieving coherent and cost-efficient long-form MV generation.

\begin{table}[!t]
\centering
\small
\setlength{\tabcolsep}{5pt}
\renewcommand{\arraystretch}{1.12}
\caption{Comparison with previous MV generation methods on the five-song pilot benchmark. BeatAlign measures music-video synchronization, CQR evaluates the quality-cost trade-off, CLIP measures visual-semantic alignment with lyrics, Motif measures long-range visual consistency, and Cost denotes the average generation cost (USD) for the evaluated excerpt generation. Best results are highlighted in bold.}
\label{tab:baseline_comparison}
\resizebox{\textwidth}{!}{%
\begin{tabular}{lccccc}
\toprule
\textbf{Method}
  & \textbf{BeatAlign}$\uparrow$
  & \textbf{CQR}$\uparrow$
  & \textbf{CLIP}$\uparrow$
  & \textbf{Motif}$\uparrow$
  & \textbf{Cost}$\downarrow$ \\
\midrule
MuseV~\cite{musev2024}          & $0.0834 \pm 0.0212$ & $0.2082 \pm 0.0286$ & $0.2517 \pm 0.0195$ & $0.4215 \pm 0.0015$ & $3.04 \pm 0.19$ \\
VideoComposer~\cite{Wang2024videocomposer}  & $0.1021 \pm 0.0242$ & $0.2210 \pm 0.0311$ & $0.2327 \pm 0.0223$ & $0.4186 \pm 0.0124$ & $3.15 \pm 0.21$ \\
AutoMV~\cite{Tang2025AutoMV}         & $0.0966 \pm 0.0237$ & $0.4702 \pm 0.0360$ & $\mathbf{0.3226 \pm 0.0175}$ & $0.4076 \pm 0.0139$ & $3.25 \pm 0.22$ \\
\midrule
\method{} (Ours)
               & $\mathbf{0.6689 \pm 0.0393}$
               & $\mathbf{0.7596 \pm 0.0348}$
               & $0.3011 \pm 0.0184$
               & $\mathbf{0.4784 \pm 0.0146}$
               & $\mathbf{2.19 \pm 0.10}$ \\
\bottomrule
\end{tabular}%
}
\end{table}

\subsection{Ablation Study}
\label{sec:component_ablation}
We further analyze the contribution of each component in \method{} by removing the allocation optimizer, beat-synchronous assembly, and motif-aware reuse while keeping the remaining generation process fixed. Results are shown in \cref{tab:component_ablation}. \method{} achieves the best BeatAlign, CQR, and motif consistency while almost maintaining the CLIP score. The slightly higher CLIP after removing motif-aware reuse (\textit{w/o Reuse}) is expected because independent regeneration better matches local semantics, whereas motif-aware reuse prioritizes long-horizon visual consistency. Such ablated setting further weakens motif consistency and cost efficiency, verifying that shared-prefix reuse preserves recurring visual structures while avoiding redundant synthesis. Removing allocation (\textit{w/o Allocation}) causes the largest degradation in CQR, demonstrating that the group-level optimizer improves budget utilization by jointly considering segment importance and motif dependencies. Removing beat-synchronous assembly (\textit{w/o Beat-Sync}) substantially reduces rhythmic alignment without affecting generation cost, confirming that beat-synchronous assembly primarily improves temporal coherence. Overall, the ablation results demonstrate that the three components address complementary challenges, with allocation improving quality-cost optimization, Beat-Sync improving rhythmic coherence, and reuse improving long-horizon visual consistency.

\begin{table}[!t]
\centering
\small
\setlength{\tabcolsep}{5pt}
\renewcommand{\arraystretch}{1.12}
\caption{Ablation of the major components on the complete-song benchmark. BeatAlign measures music-video synchronization, CQR evaluates the quality-cost trade-off, CLIP measures visual-semantic alignment with lyrics, Motif measures long-range visual consistency, and Cost denotes the average generation cost (USD) for complete-song generation. Best results are highlighted in bold.}
\label{tab:component_ablation}
\resizebox{\textwidth}{!}{%
\begin{tabular}{lccccc}
\toprule
\textbf{Variant}
  & \textbf{BeatAlign}$\uparrow$
  & \textbf{CQR}$\uparrow$
  & \textbf{CLIP}$\uparrow$
  & \textbf{Motif}$\uparrow$
  & \textbf{Cost (\$)}$\downarrow$ \\
\midrule
w/o Allocation
  & $0.8077 \pm 0.0428$
  & $0.4274 \pm 0.0376$
  & $0.2948 \pm 0.0055$
  & $0.4426 \pm 0.0151$
  & $5.3665 \pm 1.4177$ \\
w/o Beat-Sync
  & $0.3169 \pm 0.0343$
  & $0.5650 \pm 0.0319$
  & $0.2922 \pm 0.0051$
  & $0.4601 \pm 0.0157$
  & $\mathbf{5.1290 \pm 0.9236}$ \\
w/o Reuse
  & $0.8151 \pm 0.0434$
  & $0.4213 \pm 0.0233$
  & $\mathbf{0.2999 \pm 0.0045}$
  & $0.4288 \pm 0.0153$
  & $5.2991 \pm 0.9264$ \\
\midrule
\method{}
  & $\mathbf{0.8241 \pm 0.0433}$
  & $\mathbf{0.6015 \pm 0.0324}$
  & $0.2971 \pm 0.0042$
  & $\mathbf{0.4775 \pm 0.0155}$
  & $\mathbf{5.1290 \pm 0.9236}$ \\
\bottomrule
\end{tabular}%
}
\end{table}

% \textit{Uniform-Mid} assigns every independently generated segment, shared prefix, and member-specific suffix to the medium-cost generation option while retaining reuse.
% \textit{Uniform-High} assigns these components to the high-cost generation option while retaining reuse and serves as an unconstrained high-cost reference.
% \textit{Heuristic} uses a hand-designed rule to map segment saliency to generation options without solving the global allocation problem.
% This comparison tests whether the measured benefit of \method{} is associated with structured resource allocation rather than simply using a cheaper, more expensive, or locally rule-based policy.

\subsection{Qualitative Evaluation of State-Guided Coordination}
\begin{figure}[!t]
  \centering
  \safeincludegraphics[width=1\linewidth]{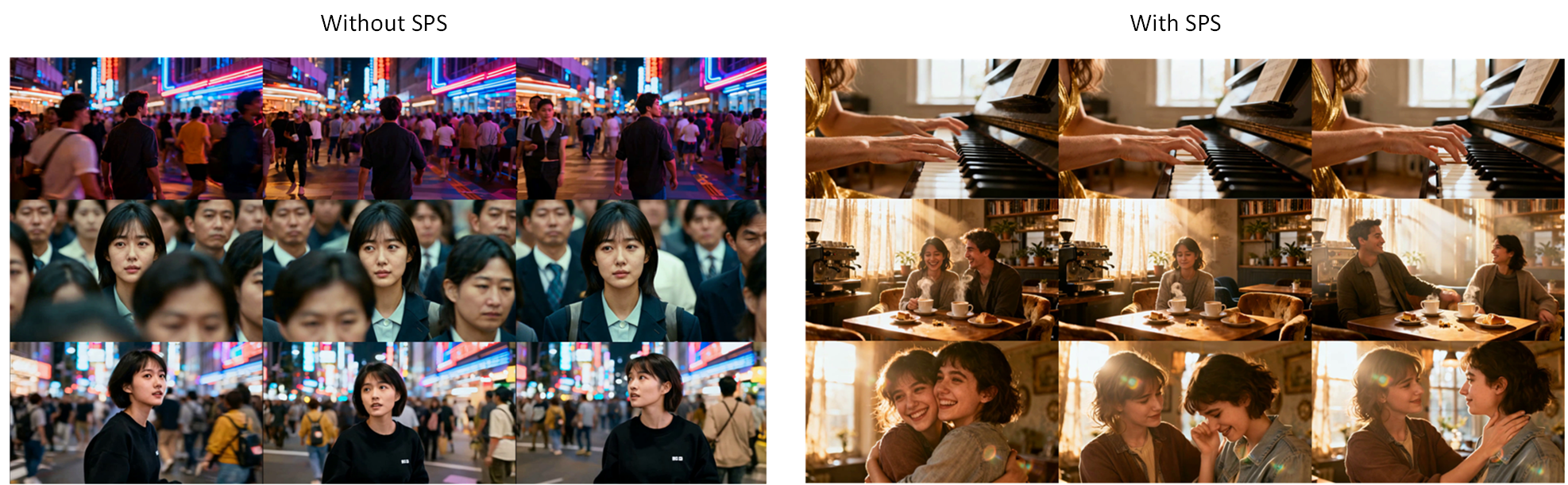}
  \caption{Qualitative comparison of character identity consistency without and with SPS during long-form music video generation. The left and right panels present results without and with SPS, respectively. In every panel, each row correspond to a different segment of the same generated video, comprising 3 frames sampled around 1-second intervals. SPS preserves persistent character references across segments, reducing identity drift during long-horizon generation.}
  \label{fig:spsl_consistency}
\end{figure}

As shown in \cref{fig:spsl_consistency}, we further examine the effect of SPS guidance on long-horizon character consistency through a qualitative comparison. Without SPS, the generated video exhibits noticeable character-identity drift, where facial attributes, clothing details, and overall appearance gradually change across segments despite describing the same character. This inconsistency becomes more apparent when recurring characters reappear after longer temporal intervals, as independent generation lacks a mechanism for maintaining shared visual references. In contrast, incorporating SPS preserves persistent character references throughout the generation process, enabling different segments to retrieve consistent visual conditions and maintain a stable character identity across both local transitions and long-range repetitions. These results demonstrate that state-guided coordination provides an effective mechanism for reducing identity drift in long-form music video generation by explicitly maintaining and propagating visual continuity information.

% Overall, this analysis shows that the proposed group-level policy achieves a stronger measured trade-off than the uniformly cheap, uniformly expensive, and locally rule-based policies on this benchmark. 
% \textit{Uniform-Mid} is cost-saving but quality-limited, \textit{Uniform-High} is expensive but inefficient, and \textit{Heuristic} uses saliency but lacks global optimization. 
% \method{} provides a better compromise by jointly considering segment importance, generation option quality, motif-sharing structure, and the global budget constraint.

\subsection{Allocation Policy Analysis}
\label{sec:allocation_policy}
We evaluate whether the proposed group-level allocation outperforms simpler budget policies on the complete-song benchmark. All policies share the same perception outputs, script plans, visual anchors, sharing graph, owner-consumer relations, motif-reuse mechanism, random seeds, and beat-synchronous assembly protocol, differing only in High/Mid action assignments. \textit{Uniform-Mid}, \textit{Uniform-High}, and \textit{Heuristic} represent medium-cost, high-cost, and saliency-based allocation strategies, respectively. As shown in \cref{tab:allocation_policy}, \method{} achieves the best BeatAlign, CQR, CLIP, and motif consistency, demonstrating improved rhythmic synchronization, visual-semantic alignment, long-horizon visual consistency, and generation efficiency. \textit{Uniform-Mid} achieves the lowest cost but sacrifices CQR, CLIP, and motif consistency, indicating under-allocation of musically or narratively important segments. \textit{Uniform-High} leads to the highest cost and the lowest CQR, while without improving the remaining evaluation metrics, indicating that uniformly applying high-cost generation is both inefficient and ineffective. Although \textit{Heuristic} exploits saliency, it still underperforms \method{} since it cannot jointly optimize global budget allocation and motif-sharing dependencies. Comparatively, the group-level MCKP planner jointly optimizes saliency-weighted utility and motif reuse under a global budget, leading to improved performance.

\begin{table}[t!]
\centering
\small
\setlength{\tabcolsep}{5pt}
\renewcommand{\arraystretch}{1.12}
\caption{Comparison of budget allocation policies on the complete-song benchmark. Uniform-Mid and Uniform-High apply fixed Mid and High generation actions to all segments, respectively, while Heuristic allocates generation quality based on segment saliency without modeling global budget competition and motif dependencies. BeatAlign measures music-video synchronization, CQR evaluates the quality-cost trade-off, CLIP measures visual-semantic alignment with lyrics, Motif measures long-range visual consistency, and Cost denotes the average generation cost (USD). Best results are highlighted in bold.}
\label{tab:allocation_policy}
\resizebox{\textwidth}{!}{%
\begin{tabular}{lccccc}
\toprule
\textbf{Policy}
  & \textbf{BeatAlign}$\uparrow$
  & \textbf{CQR}$\uparrow$
  & \textbf{CLIP}$\uparrow$
  & \textbf{Motif}$\uparrow$
  & \textbf{Cost (\$)}$\downarrow$ \\
\midrule
Uniform-Mid
  & $0.8118 \pm 0.0432$
  & $0.5144 \pm 0.0795$
  & $0.2545 \pm 0.0049$
  & $0.4136 \pm 0.0152$
  & $\mathbf{3.8030 \pm 0.7741}$ \\
Uniform-High
  & $0.8142 \pm 0.0434$
  & $0.3487 \pm 0.0269$
  & $0.2810 \pm 0.0055$
  & $0.4234 \pm 0.0154$
  & $8.0230 \pm 1.7693$ \\
Heuristic
  & $0.8100 \pm 0.0429$
  & $0.4929 \pm 0.0534$
  & $0.2873 \pm 0.0024$
  & $0.4353 \pm 0.0148$
  & $4.2121 \pm 0.9289$ \\
\midrule
\method{}
  & $\mathbf{0.8241 \pm 0.0433}$
  & $\mathbf{0.6015 \pm 0.0324}$
  & $\mathbf{0.2971 \pm 0.0042}$
  & $\mathbf{0.4775 \pm 0.0155}$
  & $5.1290 \pm 0.9236$ \\
\bottomrule
\end{tabular}%
}
\end{table}

% The previous subsections evaluate the end-to-end behavior of \method{} and the contribution of individual modules. 
% We further examine whether the planner itself is reliable and practical. 
% This analysis focuses on three questions: whether the saliency and quality signals used by the allocation objective are consistent with observed judgments, whether the motif-reuse execution contract is faithfully realized, and whether the group-level MCKP solver scales to larger planning instances.

\subsection{Planning Signal Validation}
The allocation objective relies on segment saliency $s_i$ and action-level quality proxy $Q(\cdot)$. We validate whether these planning signals provide reasonable guidance by comparing saliency scores with human judgments and calibrating the relative quality difference between High and Mid generation actions. Results show that Qwen-Plus saliency scores strongly agree with human ratings, achieving Pearson correlation of $0.936$ and Spearman correlation of $0.921$, with a mean absolute error of $0.389$ and neither a paired $t$-test nor a Wilcoxon signed-rank test detects systematic score inflation. For quality calibration, we compare realized High and Mid segments on a held-out pilot song. The observed quality ratio between the two generation options is $1.554$, with a bootstrap $95\%$ confidence interval between $1.215$ and $1.993$, covering the predefined planning ratio of $1.429$. Together with the saliency agreement results, these findings support the reliability of the planning signals for segment ranking and the calibration of relative quality differences between generation options, while the small validation sample suggests that they should be interpreted as planning proxies rather than universal perceptual-quality measures.

% \begin{table}[t]
% \centering
% \small
% \setlength{\tabcolsep}{5pt}
% \renewcommand{\arraystretch}{1.08}
% \caption{Pilot validation of the signals used by the planner.}
% \label{tab:signal_validation}
% \begin{tabular}{@{}lcc@{}}
% \toprule
% \textbf{Signal} & \textbf{Statistic} & \textbf{Value} \\
% \midrule
% \multirow{3}{*}{Saliency}
%   & Pearson / Spearman & $0.936 / 0.921$ \\
%   & MAE & $0.389$ \\
%   & paired $t$ / Wilcoxon $p$ & $0.163 / 0.153$ \\
% \midrule
% \multirow{2}{*}{Quality}
%   & High/Mid ratio & $1.554$ \\
%   & bootstrap 95\% CI & $[1.215,1.993]$ \\
% \bottomrule
% \end{tabular}
% \end{table}

\subsection{Motif-Reuse Verification.}
The motif-reuse mechanism operates at a pair level, and we therefore perform a controlled execution verification on $25$ frozen motif-pair instances. Each instance contains $5$ components: the owner's full segment, shared prefix, and the consumer's independent full segment, divergent reuse suffix, and final assembled consumer segment. The evaluation is conducted in a deterministic scene using recorded execution traces, eliminating the influence of backend API fluctuations. Results show that all $25$ pairs satisfy the verification criteria across $125$ components, with a mean prefix SSIM of $0.994$, indicating that the reused prefix is consistently preserved during retrieval and assembly. Moreover, motif reuse reduces the total generation cost from $6.21$ to $4.66$, achieving a $25.0\%$ saving compared with independent generation. These findings verify that the shared-prefix, divergent-suffix strategy faithfully executes the reuse plan and provides the expected synthesis cost reduction under recorded traces.

% ================================================================
%  5. Conclusion
% ================================================================
\section{Conclusion}
\label{sec:conclusion}

We presented \method{}, a state-guided multi-agent framework for budget-aware music video generation. By coordinating specialized generation stages through a Structured Persistent State, \method{} transforms MV generation from independent clip synthesis into a coordinated production process that maintains visual dependencies, supports motif reuse, and enables explicit resource allocation. The proposed group-level MCKP formulation further allows the system to select generation strategies under a global budget constraint, improving the balance between generation quality and synthesis cost. Experiments on song benchmarks demonstrate that \method{} achieves improved overall performance across quality, semantic alignment, rhythmic synchronization, and visual consistency metrics compared with existing MV generation frameworks, while additional analyses validate the effectiveness of model components. Future work will explore more flexible coordination strategies and larger-scale evaluation with human preference studies for diverse long-form video generation scenarios.

% \clearpage

\bibliographystyle{splncs04}
\bibliography{main}

\end{document}